\documentclass{egpubl}
\usepackage{EGUKCGVC2024}
 
\WsSubmission      

\usepackage[T1]{fontenc}
\usepackage{dfadobe}  

\usepackage{cite}  
\BibtexOrBiblatex
\electronicVersion
\PrintedOrElectronic
\ifpdf \usepackage[pdftex]{graphicx} \pdfcompresslevel=9
\else \usepackage[dvips]{graphicx} \fi

\usepackage{egweblnk} 

\title[Semantic UV mapping to improve texture inpainting for indoor scenes]%
      {Semantic UV mapping to improve texture inpainting for indoor scenes}

\author{
 Jelle Vermandere, 
 Maarten Bassier,
 Maarten Vergauwen
}

\begin{document}


\maketitle
\begin{abstract}
  This work aims to improve texture inpainting after clutter removal in scanned indoor meshes. This is achieved with a new UV mapping pre-processing step which leverages semantic information of indoor scenes to more accurately match the UV islands with the 3D representation of distinct structural elements like walls and floors. Semantic UV Mapping enriches classic UV unwrapping algorithms by not only relying on geometric features but also visual features originating from the present texture. The segmentation improves the UV mapping and simultaneously simplifies the 3D geometric reconstruction of the scene after the removal of loose objects. Each segmented element can be reconstructed separately using the boundary conditions of the adjacent elements. Because this is performed as a pre-processing step, other specialized methods for geometric and texture reconstruction can be used in the future to improve the results even further.


\end{abstract}  
\section{Introduction}
Empty 3D indoor environments, captured from real locations, are in high demand in the gaming and Architecture, Engineering, Construction, and Operations (AECO) industries~\cite{Vermandere2022}. These environments can be used for a wide variety of applications, such as remodeling, renovations, and interactive simulations.
These environments can be captured and processed using different methods. One of the more popular methods involves using a 3D scanner to capture a full 3D point cloud accompanied by panoramic images that add more information. For efficient consumption, these models are converted to meshes, which retain much of the geometric detail while also embedding the textural information of the surfaces \cite{Bassier2024}.
However, these environments are rarely completely empty when captured. Loose objects present during the capture process can lead to occlusions, either because they are placed against a permanent element or because they block the view of another part of the room. Current semantic instance segmentation methods can automatically detect these objects \cite{Dai2018}, enabling an automated removal process. Removing these objects from the scene reveals occlusions and holes, resulting in an incomplete environment. Therefore, there is a need to complete these missing parts.

Holes and missing regions in meshes can be completed in two steps: first geometrically and second texturally. 
Geometric hole filling has been a field of much research which has lead to very robust tools and algorithms \cite{Dai2018, Mittal2022, Boissonnat2014} to fill these holes.
Image inpainting has seen a recent increase in popularity due to the rise of diffusion models which dramatically improve inpainting results \cite{lugmayrRepaint}.
However, there are still some obstacles to use this method on 3D model textures. The visual look of a 3D object is created by using a texture map which projects the faces on a 2D plane. This projection is called UV projection. The projection process creates a certain disconnect between the 3D mesh and the UV texturemap \cite{Vermandere2023}. the relation between the location of a face in 3D is not necessarily the same on the UV plane. This means adjacent faces do not always remain adjacent in 2D.

Current SOTA works approach this problem in different ways. 
Works like \cite{Gkitsas2021,Slavcheva2024, Wei2023} use a 2D inpainting approach to paint on the camera views and reproject them on the mesh, while this works well for objects close to walls or distant from the camera, these methods have difficulty handling large occlusions due to complex room geometry of multiple objects covering the views.
Other works \cite{Flynn2022, Oechsle2019} aim to directly predict the color in 3D space. These models are however limited in resolution due to the use of vertex colors or texture fields.

The main goal of this work is to improve the uv projection of the scene by leveraging the semantic instance segmentation to separate the loose parts from the scene as well as the different structural elements like walls and floor.
Using these masks the loose objects can be removed from the scene and the resulting missing geometry can be reconstructed element by element. Furthermore, the instanced structural elements also allow for a better uv mapping, ensuring the resulting uv map closer matches the 3D mesh, both to minimise distortion and keep the adjacent faces together. This new uv map will improve the texture inpainting process and create a better looking mesh.

\begin{figure*}[!h]
    \centering
    \includegraphics[width=\textwidth]{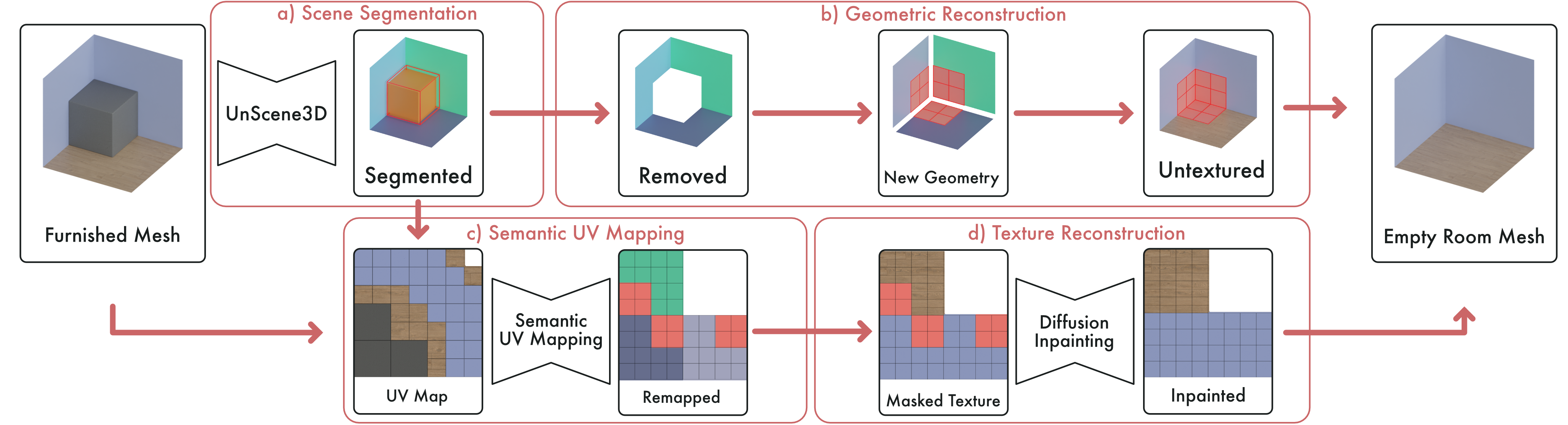}
    \caption{Overview of the proposed pipeline, starting with a furnished mesh (left), featuring the parallel scene segmentation and geometric reconstruction (top), and the semantic UV mapping and texture reconstruction (bottom) to result in an empty room mesh (right).}
    \label{fig:methodology}
\end{figure*}

\section{Background and related work}
\label{sec:background}

\subsection{Texture inpainting}
Restoring missing parts of an image has evolved from algorithm-based works like Gaussian inpainting \cite{Galerne2017} to machine learning based approaches like Inpaint anything \cite{Yu2023} and Mask-based inpainting \cite{Li2022}. These have the advantage being able to predict the missing pixels based on the surrounding data instead of solely extrapolating a pattern from the image.

The shift towards diffusion based inpainting has enabled works like No shadow left behind \cite{Zhang2021} to removed masked object completely from a picture, including the object's shadows. PanoDR \cite{Gkitsas2021} and work that build on it \cite{Slavcheva2024} take this a step further and train the diffusion model on spherical panoramic images to enable direct object removal on 360 images. Because these models operate purely in 2D they do not contain any 3D representation of the scene.

Diffusion based inpainting has also been used to remove objects from a scene in 3D. Clutter detection and removal \cite{Wei2023} inpaints both RGB and Depth images from multiple viewpoints of a single object and reconstructs the 3D mesh in those missing parts. NeRFiller \cite{Weber2023} uses a similar approach but creates a Neural Radiance Field (NeRF) instead. These view based models are limited by was is visible by the camera in a single view. Instead of using a camera view of the missing region, Texture inpainting for photographic models \cite{Maggiordomo2023} uses dynamic UV mapping to ensure the missing region is always centered and surrounded by reference pixels to perform the inpainting, but is limited to small areas.

The missing color can also be predicted directly on the mesh. STINet \cite{Flynn2022} directly predicts the vertex colors of the missing regions. While Texture fields \cite{Chen2022} creates an implicit neural field to generate the missing regions. These methods however are limited by the resolution of the geometry and struggle to generate fine details.

\subsection{Scene Segmentation}

When trying to segment a scene, the different objects need to be detected. Works like Votenet \cite{Ding2020} or V detr \cite{Shen2023} use a point-transformer-model \cite{Wu2023} to create bounding boxes for each distinct object. While these work well, they only detect objects.
Full scene instance segmentation takes this a step further and labels every face. Works like Unscene3d \cite{Rozenberszki2023} Can perform a class-agnostic segmentation completely unsupervised. Sai3D \cite{Yin2023} Also enables CLIP based embedding to search for specific objects in the scene.

\subsection{UV mapping}

The biggest obstacle in using 2D inpainting methods on 3D meshes is the lack of a UV map that is both efficient and retains the face adjacent relations of objects in a scene. Graphseam \cite{Teimury2020} uses a Graph Neural Network (GNN) to automate the UV mapping process while retaining semantic seams, while flatten anything \cite{Zhang2024} uses point-wise mappings between the 3D points and UV coordinats. These methods are however difficult to generalise to a large scene. Nuvo \cite{Srinivasan2023} aims to combat this by optimizing the UV layout for the visible parts by using a neural field. This largely overcomes the struggles of the complex geometry from reconstructed scenes.

\section{Methodology}
\label{sec:methodology}
The proposed method as shown in Figure \ref{fig:methodology} consists of multiple steps: First the input mesh is segmented, the segmentation masks are used for both element separation and UV seam creation. Second, The loose objects are removed and the segmented structural elements are all completed geometrically. Third, the UV map is unwrapped following the semantic seams. Fourth, the texture is inpainted in the newly created geometry. Finally, the texture is reprojected on the empty mesh.

\subsection{Scene segmentation}
The first step is segmenting the full scene as seen in Figure \ref{fig:methodology}a. Before the segmentation is performed, due to the limited resolution of the mesh, we cannot guarantee that each face is exclusive to an single object. This is why we first perform a Geometry refinement step \cite{Vermandere2023} to split the faces according to there texture. Both the loose objects and the structural elements are detected using UnScene3D \cite{Rozenberszki2023} Which uses both geometric and color features to generate pseudomasks, these masks are then refined using a self trained model. Since the model is optimised for object detection, the structural elements like walls can sometimes remain clustered together. This is why we also perform a RANSAC plane segmentation \cite{Korman2018} to refine the walls.

\subsection{Geometric Reconstruction}
The loose objects, detected in the previous step, are removed from the scene as illustrated in Figure \ref{fig:methodology}b. This results in large holes that need to be reconstructed. Before each segmented structural element is reconstructed one-by-one, the RANSAC planes, detected in the previous step, are used to determine the intersection edges between the elements. These edges form the boundary conditions for the Delaunay reconstruction\cite{Boissonnat2014}. Depending on the number of intersecting planes different boundary faces are created according to Figure \ref{fig:geometryReconstruction}.

\begin{figure}[!h]
    \centering
    \includegraphics[width=\columnwidth]{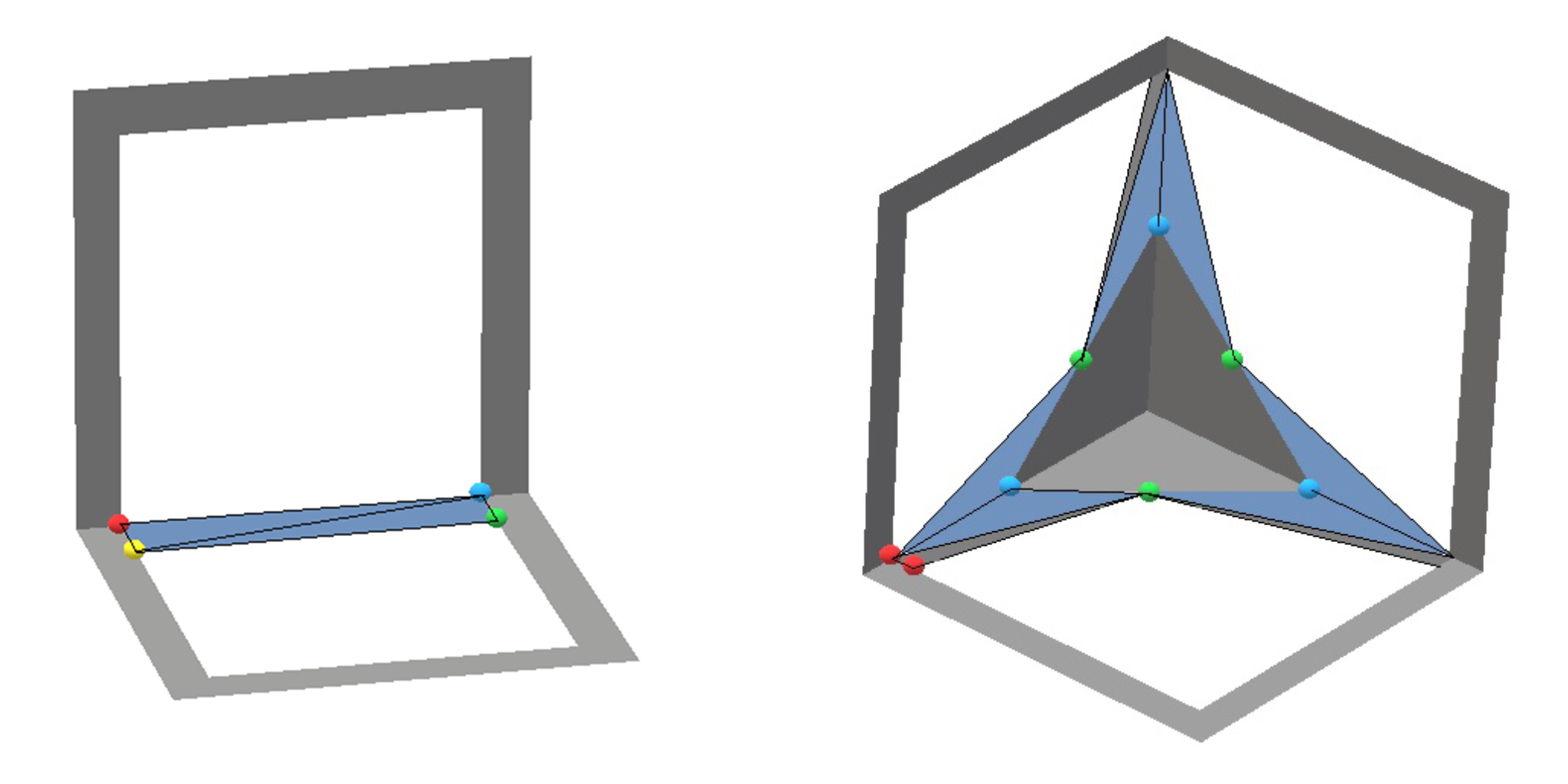}
    \caption{The two-plane-intersection case (left) and the three-plane-intersection case (right)}
    \label{fig:geometryReconstruction}
\end{figure}

\subsection{Semantic UV Mapping}
After the geometry has been reconstructed, the newly generated faces are given the same label as the original element. The boundaries between the different segmentation labels are marked as UV seams. These seams serve as the basis for the semantic UV mapping as seen in Figure \ref{fig:methodology}c. Since the end goal is to inpaint the missing areas, we optimize the UV map to minimize distortion, while still aiming to keep as many faces as possible from the same label joined. This is largely possible due to the flat nature of the structural elements in indoor scenes. Furthermore, when inpainting textures the relevant orientation of the image is also relevant. By introducing a Y/Z up consistency in our unwrapping method, each element is oriented consistently. where vertical elements like walls and beams are always oriented with the up-direction facing up on the image, flat elements like floors and ceilings have their forward direction facing up.

\subsection{Texture reconstruction}
The final step after the mesh has been semantically UV mapped is painting in the missing regions. This is performed on the 2D texture of each element. The newly generated geometry serves as the inpainting mask, this ensures only the new parts are altered. The rest of the UV island serves as reference for the diffusion based inpainting \cite{lugmayrRepaint}. Because each element is inpainted separately, there is no confusion from other adjacent materials possible.

After the Texture has been inpainted completely. The texture is reprojected on the 3D mesh. Because the UV map was optimized for inpainting, and not for efficiency, the resulting UV maps can be very large. This is why, for a final step we repack the UV layout for optimized space efficiency, while keeping the semantic islands in tact.

\section{Experiments}
\label{sec:experiments}

\subsection{Dataset}
\begin{figure}[!h]
    \centering
    \includegraphics[width=\columnwidth]{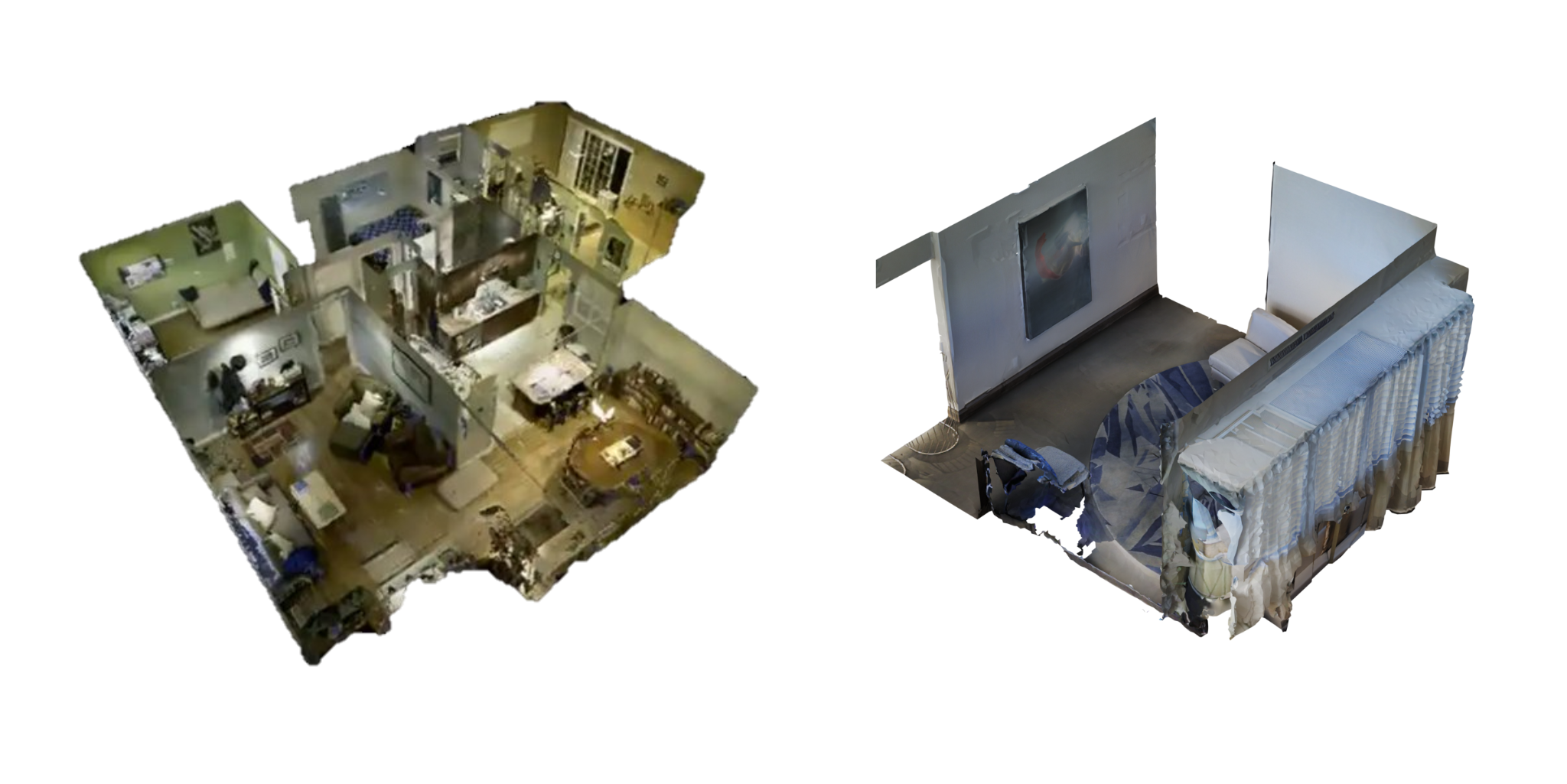}
    \caption{A scene from the ScanNet++ dataset (left) and Matterport3D dataset (right)}
    \label{fig:datasets}
\end{figure}

For our experiments, we used the ScanNet++ \cite{Yeshwanth2023} and Matterport 3D \cite{Chang2017} Datasets as seen in Figure \ref{fig:datasets}. The ScanNet++ dataset contains 460 high-resolution 3D reconstructions of indoor scenes with dense semantic and instance annotations. The Matterport 3D Dataset is a scanned dataset that consists of 90 fully textured building-scale scenes, inclusing semantic labels of the whole dataset. We focused on single room scenes with moderately dense furniture, pre-labeled. These labels serve as the baseline for both the loose object removal and the semantic UV mapping.

\subsection{Object detection and removal}
\begin{figure}[!h]
    \centering
    \includegraphics[width=\columnwidth]{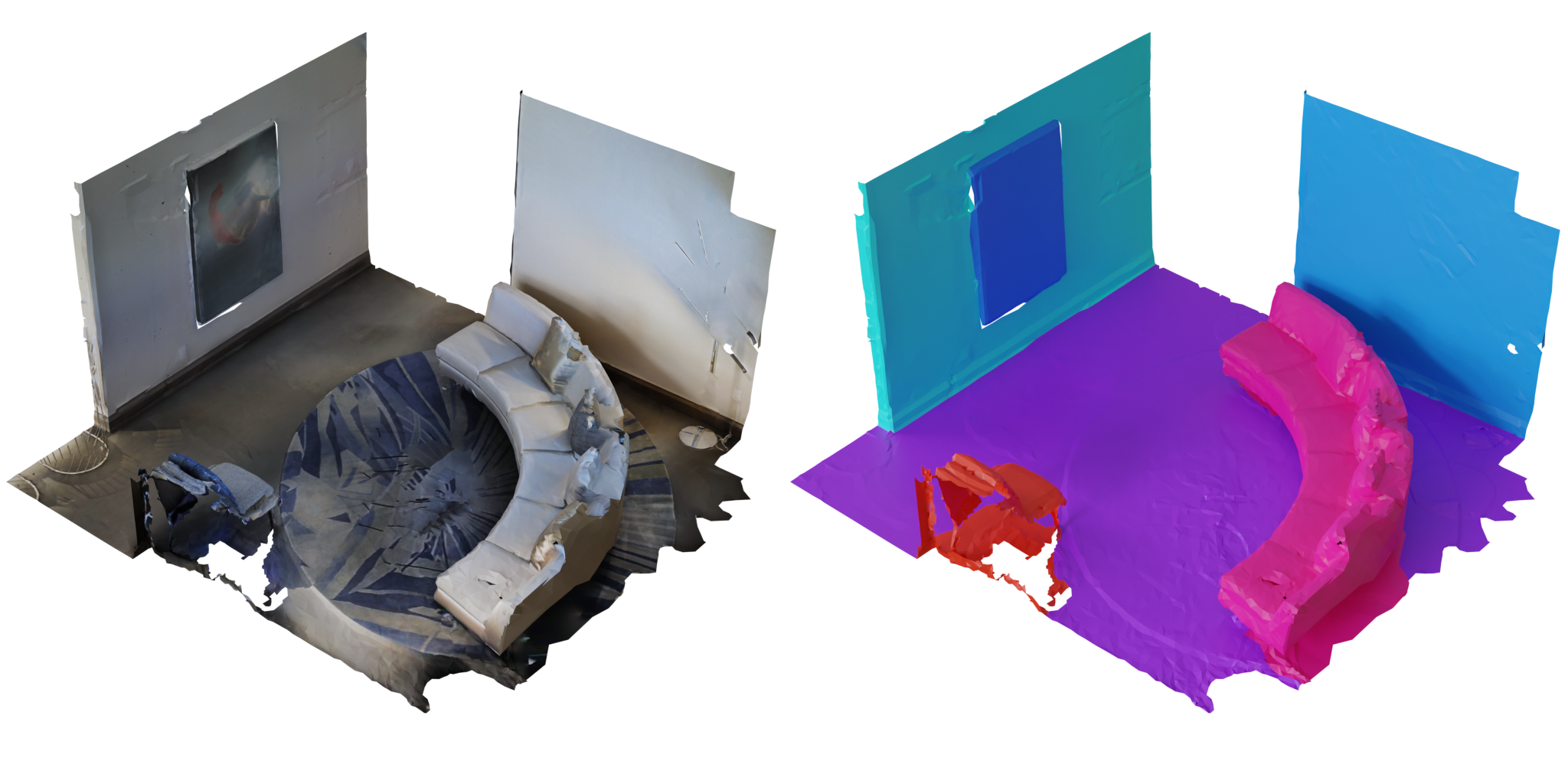}
    \caption{The matterport room(left) and the segmented labels (right)}
    \label{fig:segmentation}
\end{figure}

For our experiments, the instance masks from the Matterport3D dataset are used to separate the mesh as seen in Figure \ref{fig:segmentation}. The labels do not always align perfectly with the objects, this is why we removed all the faces inside the bounding box of the objects. this ensured a clean cut line. The RANSAC plane segmentation performed well on the walls and floors, but had difficulty with more complex geometry 

The experiments have shown that the geometry completion performs better when each object is removed sequentially, rather than in parallel. This is illustrated in Figure \ref{fig:geometryReconstruction}. Furthermore, overlapping objects can disrupt the planar detection, so they should be removed together, while this leads to a higher amount of existing data that is removed, the final results will be better.

\begin{figure}[!h]
    \centering
    \includegraphics[width=\columnwidth]{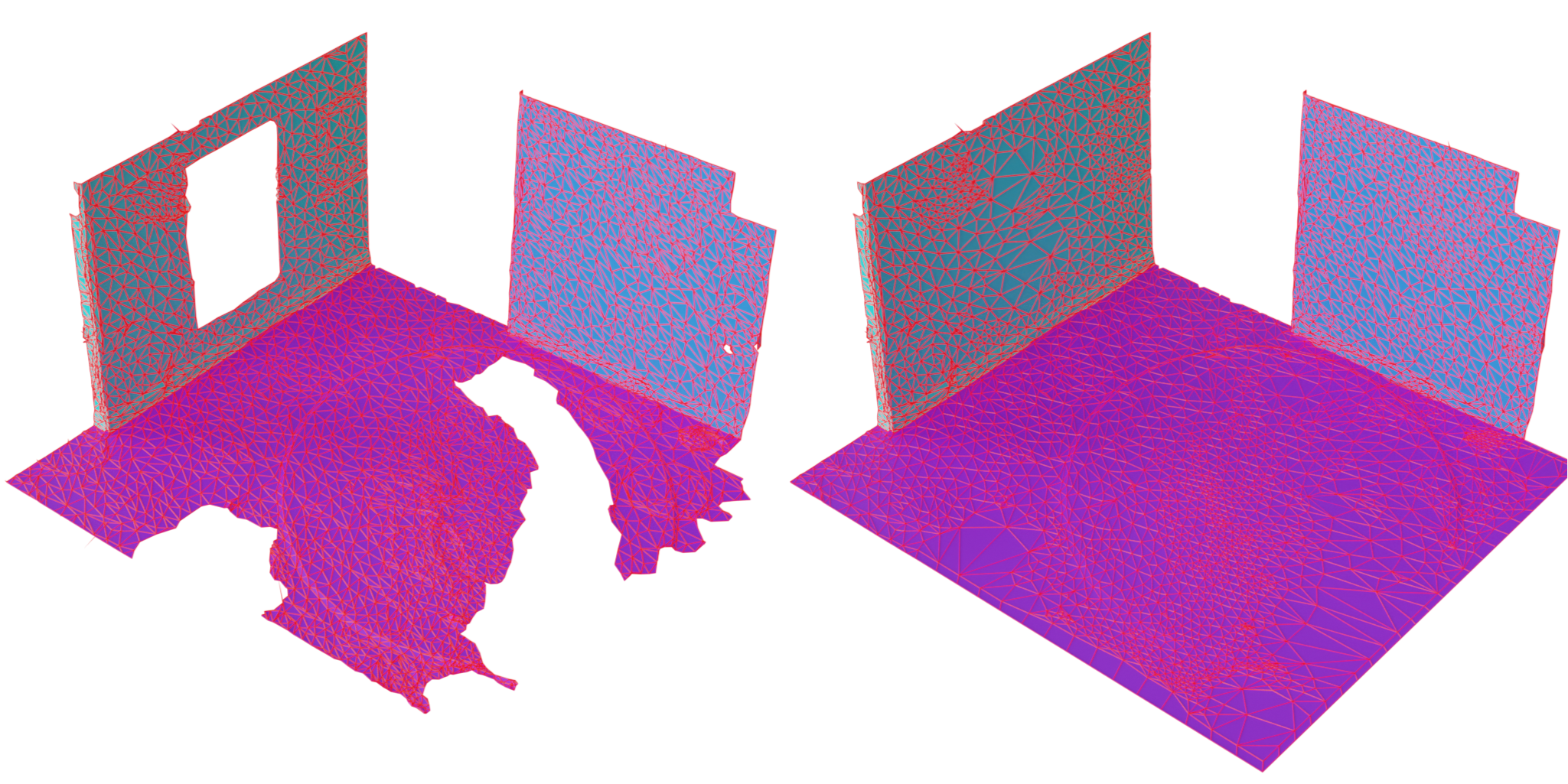}
    \caption{The object-removed room (left) and the reconstructed room (right)}
    \label{fig:geometryReconstruction}
\end{figure}

\subsection{Semantic UV Mapping}
\begin{figure}[!h]
    \centering
    \includegraphics[width=\columnwidth]{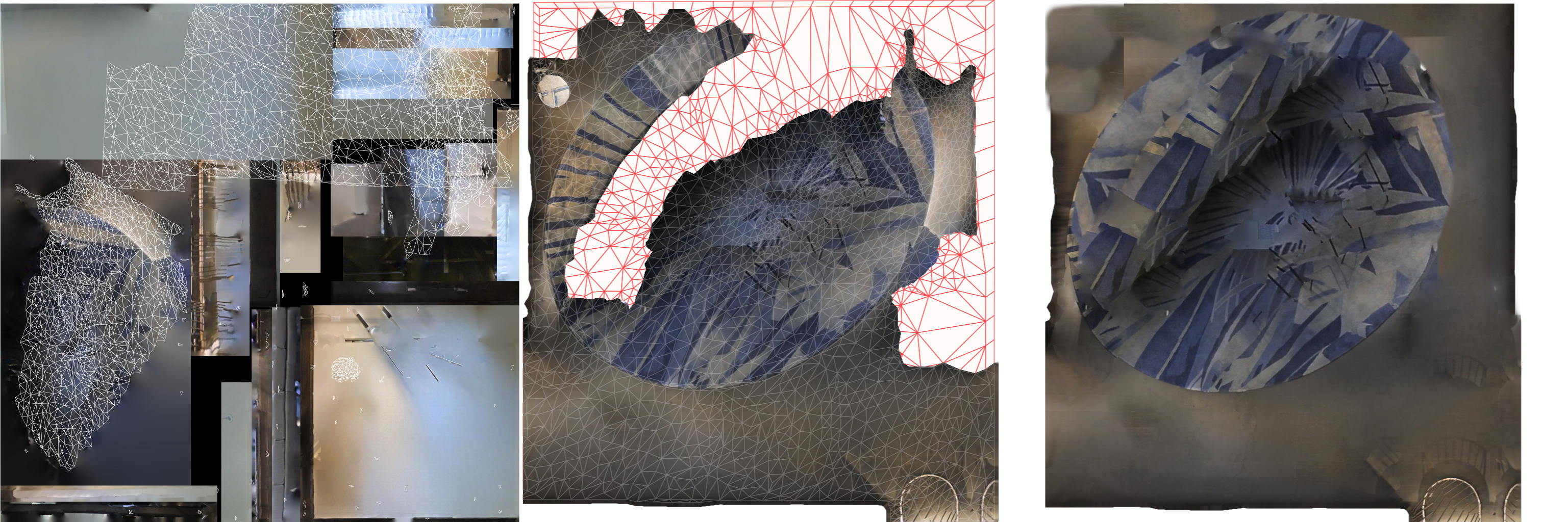}
    \caption{The original texture map (left) the new semantic UV lyout for the floor (center) and the inpainted texture (right)}
    \label{fig:uvmapping}
\end{figure}

The semantic segmentation has created the UV seams at not just geometrically distinct edges, but also texturally (Figure \ref{fig:geometryReconstruction}). Due to the simple geometry of the structural elements the uv maps can be created without to much distortion as seen in Figure Figure \ref{fig:uvmapping}. We do see however, that due to the orientation constraint, the UV maps are layed out seperatly for each object. This means the texturesize is very large, larger than the original texturemap from the dataset.

\subsection{Texture reconstruction}
\begin{figure}[!h]
    \centering
    \includegraphics[width=\columnwidth]{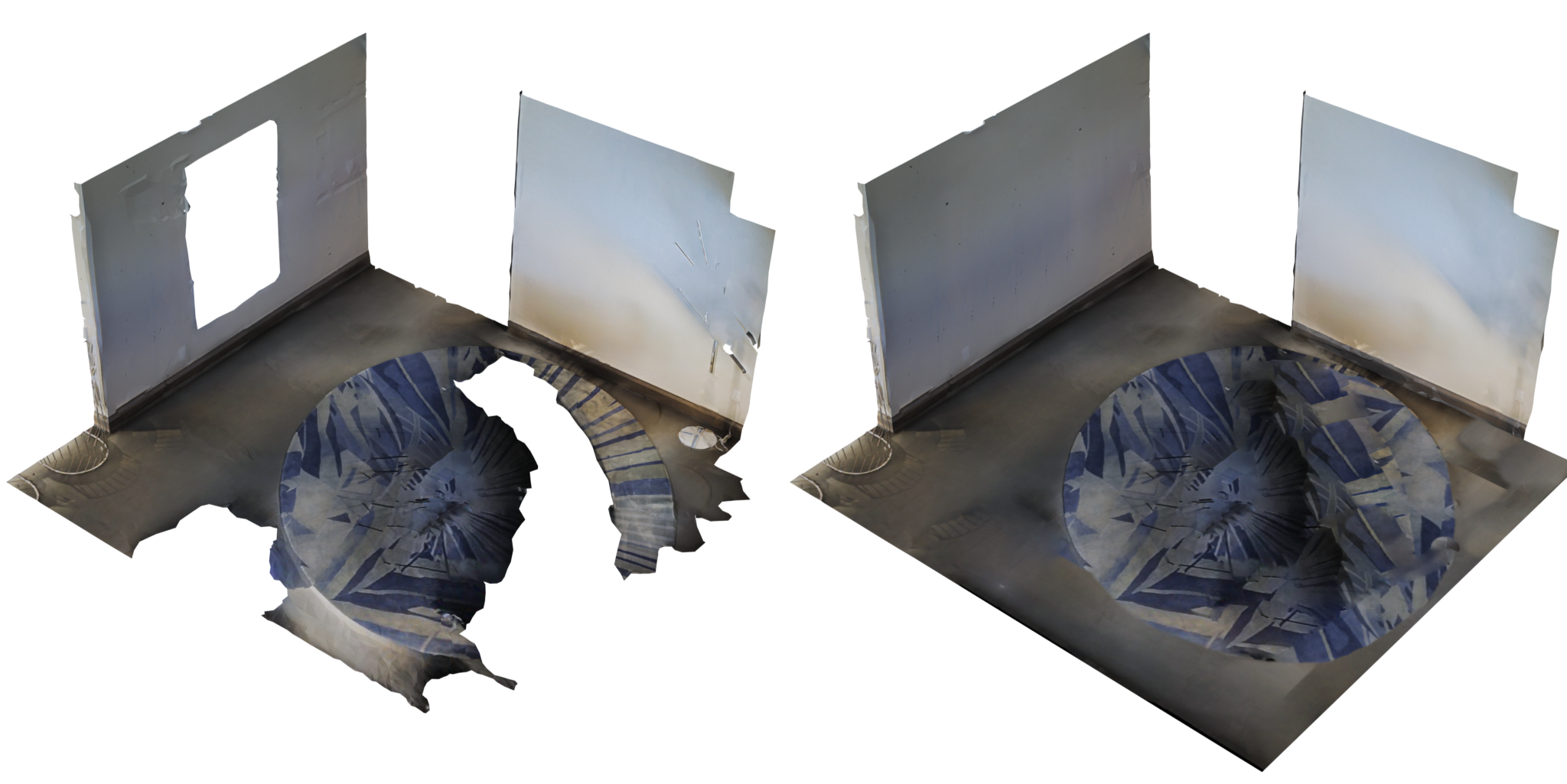}
    \caption{The object removed textured room (left) final unfurnished room (right)}
    \label{fig:result}
\end{figure}

The texture inpainting process is able to use the existing texture as an example which creates mostly indistinguishable textures for the more basic surfaces (\ref{fig:uvmapping}). The faces of the new geometry provide a clear bounding mask, allowing the inpainting to only affect the required area. The final result can be seen in Figure \ref{fig:result}

\section{Discussion}
\label{sec:discussion}

The resulting empty scenes after the loose objects have been removed show believable results. This is helped by the fact that the structural elements in indoor scenes are generally very simple. By reducing the geometric reconstruction to a planar triangulation, the problem becomes much less complex and manageable. This is however not possible for all types of structural elements. More organic shapes require a more complex Reconstruction like AUTO-SDF \cite{Mittal2022}. The advantage of our pre-processing pipeline is that the methods are interchangeable, while still retaining the advantages of the semantic segmentation.
The inpainted textures show very good results for the repetitive of basic materials. However, more graphic elements that are not properly segmented can lead to artifacting in the final results.
The effectiveness of this method is however difficult to quantify due to the lack of real ground truth data. This is why we visually evaluated each scene, checking for visual consistency and believability.

\section{Conclusion}
\label{sec:conclusion}
This paper introduced a novel pre-processing step in the object removal pipeline for indoor scanned environments. By semantically labeling the different elements in the scene, both the geometry completion and texture reconstruction can be improved due to clearer boundaries between the different elements. The holes resulting from the removal of the detected loose objects can be better completed element wise, rather than for the whole scene. Using the predicted intersectionlines between the different elements, we can clearly define the boundary conditions for the geometric Reconstruction. 
The semantic UV mapping also ensures each element is mapped as close a spossible to its 3D representation, making the inpainting process much more straightforward. The existing, textured parts of the elements serve as reference for the newly created geometry.

\newpage
\bibliographystyle{eg-alpha-doi} 
\bibliography{export}       


\end{document}